\documentclass{article}

\PassOptionsToPackage{numbers}{natbib}

\usepackage[final]{neurips_2022}

\usepackage{amsthm}         
\usepackage[utf8]{inputenc} 
\usepackage[T1]{fontenc}    
\usepackage{hyperref}       
\hypersetup{colorlinks,allcolors=blue}
\usepackage{url}            
\usepackage{booktabs}       
\usepackage{amsfonts}       
\usepackage{nicefrac}       
\usepackage{microtype}      
\usepackage{xcolor}         
\usepackage{enumitem}       
\usepackage{framed}         

\usepackage{array}
\newcolumntype{P}[1]{>{\centering\arraybackslash}p{#1}}
\usepackage{multirow}

\usepackage{xspace}
\usepackage{wrapfig}
\usepackage{graphicx}
\usepackage{MnSymbol}
\usepackage{ifthen}
\usepackage{cleveref}
\newboolean{include-notes}
\setboolean{include-notes}{true}

\usepackage{algorithm}
\usepackage{algpseudocode}

\usepackage{amsmath}
\usepackage{amsfonts}
\DeclareMathOperator*{\argmax}{arg\,max}

\usepackage[T1]{fontenc}

\usepackage{graphicx}
\usepackage{subcaption}
\usepackage{calligra}
\usepackage{dblfloatfix}
\usepackage{algorithm}
\usepackage{algpseudocode}

\usepackage{hyperref}
\usepackage{ifthen}
\usepackage{xcolor}
\usepackage[normalem]{ulem}

\newcommand{\MesutComment}[1]{\ifthenelse{\boolean{include-notes}}
 {{\color{teal}Y: #1}}{}}

\newcommand{\MicahComment}[1]{\ifthenelse{\boolean{include-notes}}
 {{\color{cyan}M: #1}}{}}

\newcommand{\adnote}[1]{\ifthenelse{\boolean{include-notes}}
 {{\color{blue}AD: #1}}{}}
 \newcommand{\adremove}[1]{\ifthenelse{\boolean{include-notes}}
 {{\color{blue}\sout{#1}}}{}}

\newcommand{\prg}[1]{\textbf{#1}}

\newcommand{\supref}[2]{\hyperref[#1]{Appendix #2}}

\newcommand{\apfigref}[1]{Appendix Figure~\ref{#1}}
\newcommand{\aptabref}[1]{Appendix Table~\ref{#1}}

\newcommand{\envdistr}{\mathcal{E}}
\newcommand{\sampleenv}{e}
\newcommand{\testlayout}{e^{\text{test}}}
\newcommand{\humandata}{D_{HH}}

\newcommand{\acr}{\text{OBP}}
\newcommand{\single}{1}
\newcommand{\multi}{\envdistr}

\newcommand{\BR}{\text{BR}}
\newcommand{\BRa}{\widehat{\text{BR}}}

\newcommand{\mlsp}{\text{SP}^\multi}
\newcommand{\mlbc}{\text{BC}^\multi}
\newcommand{\mlbcmlsp}{\text{BC}^\multi_{\acr}}
\newcommand{\mlbcmlspfreeze}{\text{BC}^\multi_{\acr+f}}

\newcommand{\slsp}{\text{SP}^\single}
\newcommand{\slbc}{\text{BC}^\single}
\newcommand{\slbcslsp}{\text{BC}^\single_{\acr}}

\title{Optimal Behavior Prior: \\Data-Efficient Human Models\\ for Improved Human-AI Collaboration}

\author{%
  Mesut Yang\thanks{Work completed while at UC Berkeley.}\\
  Amazon Search Science and AI\\
  \And
  Micah Carroll\\
  UC Berkeley \\
  \And
  Anca Dragan \\
  UC Berkeley \\
}

\begin{document}

\maketitle

\begin{abstract}
AI agents designed to collaborate with people benefit from models that enable them to anticipate human behavior. However, realistic models tend to require vast amounts of human data, which is often hard to collect. A good prior or initialization could make for more data-efficient training, but what makes for a good prior on human behavior? Our work leverages a very simple assumption: people generally act closer to optimal than to random chance. We show that using optimal behavior \emph{as a prior} for human models makes these models vastly more data-efficient and able to generalize to new environments. Our intuition is that such a prior enables the training to focus one's precious real-world data on capturing the subtle nuances of human suboptimality, instead of on the basics of how to do the task in the first place. We also show that using these improved human models often leads to better human-AI collaboration performance compared to using models based on real human data alone.
\end{abstract}

\section{Introduction}

One of the ultimate goals of the field of AI is creating agents that are able to \textit{collaborate with us} and help us achieve our goals. Such agents benefit from predictive models of human behavior -- even if those models are simply used to simulate human behavior when training an RL policy \citep{caroll2019utility}. Given the nuances and complexities of human behavior -- our decisions being riddled with hundreds of identified cognitive biases \citep{barnes1984cognitive, kahneman1982judgment} -- models of human behavior need to be grounded in real data.

At the same time, basing models solely on real human data has proven challenging. Human models tend to underperform relative to real humans \cite{McIlroyYoung2020AligningSA} and fail to generalize to unseen or uncommon situations \cite{Roelofs2022CausalAgentsAR}. The generalization problem is exacerbated in collaboration, where using RL to compute an agent policy that is the best-response to the human model (i.e. the policy that is optimal for the AI agent given that the human will act according to the policy captured by the model) will create incentives for the RL system to exploit biases in the human model by exploring states during training that are from a different distribution than those induced by human-human collaboration.

Thus, the limited human data we have is very valuable, and should be used in the most efficient manner possible to learn nuances of human behavior. One way to do so might be to use low-capacity models of human behavior with high inductive bias, e.g. which leverage Theory of Mind \citep{tom0, tom, toomanycooks, paulknott}. However, the quality of such models crucially depends on being able to add the correct inductive biases, and require extensive engineering effort. In contrast, we seek a solution that retains the expressivity of high-capacity models while addressing the data requirements from above.

\begin{figure*} [t]
\vskip -1.5em
\centering
\includegraphics[width=1\textwidth]{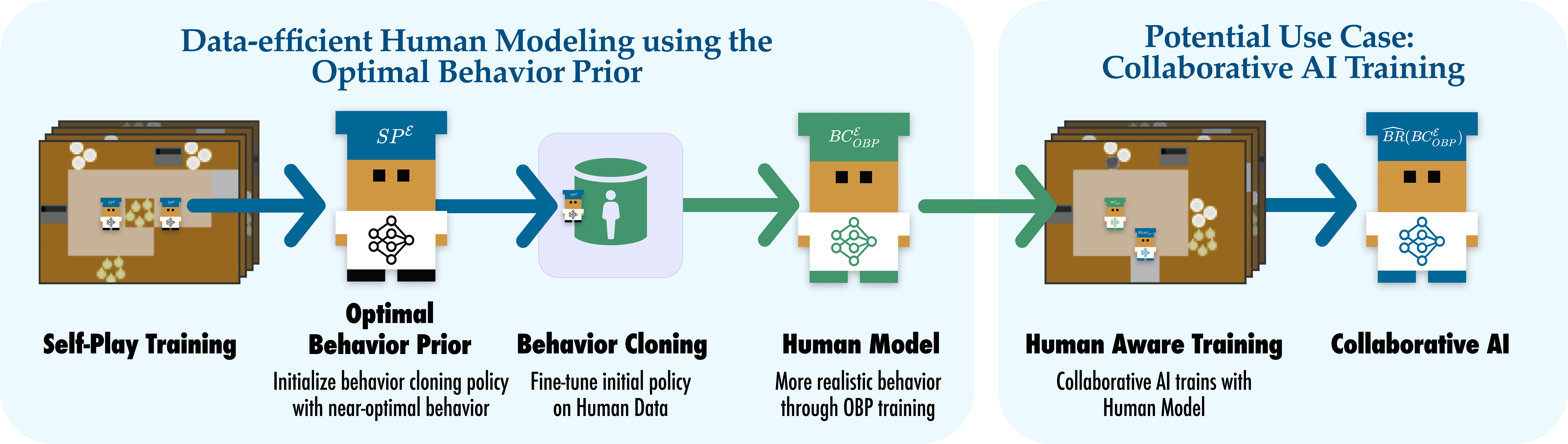}
\caption{\textbf{The Optimal Behavior Prior.} By training an agent in self-play across many different environments, one can obtain a parameterization for optimal behavior which can be reused as an initialization (or equivalently, a prior of near-optimality) for human behavior modeling. This vastly improves human modeling data-efficiency relative to a random initialization, and can also help in training collaborative AI systems which are trained with said human models, enabling generalization to unseen environments.}
\vskip -1em
\label{fig:method-overview}
\end{figure*}

Can we provide some structure for the model that, although useful in increasing the efficiency in our usage of human data, doesn't reduce our model expressivity, i.e. its ability to improve to arbitrary levels of performance as the amount of data and the coverage increase? We hypothesize that one area with high potential for improvement is providing high-capacity human models with an \textit{informed prior} about what human behavior should look like: generally, human models' prior over human behavior (corresponding to their initialization) is random. But what makes for a good prior of human behavior? 
Our insight is that \emph{while human behavior deviates from optimality in nuanced ways which are hard-to-formalize, we should leverage the fact that people are at least trying to perform tasks optimally, and use optimal behavior as a prior.} 

This builds off of the intuition that for most human tasks, people's behavior is closer to being optimal than to completely random behavior. Thus, this kind of approach will be all the more effective for settings in which human behavior is closer to optimal. But even if human behavior was not very optimal, we expect it to be much more sample-efficient to reduce a optimal model's performance down to human-level (through human-data fine-tuning), rather than to learn to increase a random model's performance up to being human-level. This kind of approach has two main advantages.

Firstly, it provides the model with a good starting representation for human behavior in the task: rather than wasting precious real human data on learning to represent basic aspects of the task (the prior takes care of that already) the learner can leverage the data for capturing the subtle nuances of human's suboptimal behavior -- following a similar idea behind pre-training and fine-tuning paradigms common in domains such as computer vision and NLP \citep{Razavian2014CNNFO, Howard2018UniversalLM}.

Additionally, although our approach requires the extra component of access to optimal behavior, it is generally easier to come by than additional human data, which would otherwise be required for good human models: self-play methods \citep{hernandez} have been shown to work surprisingly well in obtaining near-optimal policies even in complex domains \citep{alphago, Silver2017MasteringCA, alphastarblog, OpenAIFive}. To encode such an Optimal Behavior Prior (which we refer to as $\acr$), one can use the parameterization for near-optimal agents as a starting point for training one's human model.

We test this idea in Overcooked-AI \citep{caroll2019utility}, a human-AI collaboration benchmark environment based on a video game by the same name \citep{Overcooked}. In a simulated user study, we compare how different ways of modeling humans can affect a human models' quality. We find that using $\acr$ for our human models leads to qualitatively more realistic human behavior and much higher competence, enabling for the first time to obtain human models that generalize across different environments without being trained on specialized hard-coded state-featurizations. Additionally, we investigate whether such improved human model training regime can also improve downstream collaborative AIs trained using our human models. Again, we find that traditional behavior cloning models are not sufficiently realistic (in our data regime) to lead to good downstream collaborative AIs: collaborative agents obtained with simple behavior cloning perform even worse than self-play agents when paired with simulated humans. However, when using human models obtained with $\acr$ for AI training, one can achieve better collaboration performance than under all other conditions.

Overall, our results show that $\acr$ vastly increases human modeling data efficiency, suggesting that the technique could be widely applicable in any low-human-data human-modeling setting.

\section{Related work}

\prg{Human modeling.}
There has been a recent wave of interest in obtaining good human models for games \cite{McIlroyYoung2020AligningSA, Jacob2022ModelingSA, Abramson2021CreatingMI}. One of the simplest approaches to imitation learning is given by behavior cloning, which learns a policy from expert demonstrations by directly learning a mapping from observations to actions using standard supervised learning methods ~\citep{BehavioralCloning, torabi2018behavioral}. This kind of approach has been historically the most successful at imitating human behavior \citep{mcilroy-young_learning_2020, Cazenave2018ResidualNF, Silver2017MasteringTG}. However, one common problem with methods like behavior cloning is that such models tend to underperform relative to real humans~\cite{McIlroyYoung2020AligningSA}, in part likely due to insufficient expert data.
Recently, \cite{Jacob2022ModelingSA} was able to obtain more realistic human models by leveraging search to improve the human model performance, while using a behavior cloning policy regularization to keep the policy human-like. We see our approach as at least complementary to theirs: OBP models would likely constitute better human models for regularization with their method. However, OBP human models might even be competitive in isolation when compared to search-based human models, as the former will have better performance relative to traditional behavior cloning methods, potentially obviating the need for search.

\prg{Human-AI collaboration.} In this paper, we particularly focus on the benefit that improving human models can have on improving \textit{Human-AI collaboration}: building agents that can collaborate with humans \citep{caroll2019utility, human-robot-cross-training}. This is related to the \textit{ad-hoc coordination} problem -- coordination with unknown teammates \citep{stone2010, Albrecht2013AGM} -- but differs from it in that we assume that the teammate will be human. While it has been claimed that one might be able to obviate the need for human data to collaborate with humans \citep{Strouse2021CollaboratingWH}, this will not be sufficient in settings in which capturing nuances of human suboptimal behavior is essential for the tasks at hand. The knowledge that the partner will be a human can either be encoded directly through handcrafted inductive biases \citep{otherplay, toomanycooks, Hu2020SAD, Strouse2021CollaboratingWH, puig_watch-and-help_2020}, or by leveraging human data to varying degrees. Some approaches use both human data and inductive biases: first identifying which Nash equilibrium humans tend to play, and then biasing agents trained in self-play to learn the same equilibrium~\citep{lerer2019learning,Tucker2020AdversariallyGS}. However, these approaches require making strong assumptions about what test-time human behavior will look like, which may not hold in practice. This has led to various approaches to achieving human-AI collaboration which rely on learning models of human behavior and computing best-responses to them \citep{caroll2019utility, Hu2022HumanAICV}. Our approach is closest to \cite{caroll2019utility}: we first train a human model to capture human behavior, and then train a best response to it to obtain a collaborative agent. Despite requiring more human data, this enables to make no assumption about the form of human behavior (as it is learned from data). In this work, we try to make access to vast quantities of human data less of a bottleneck, both for quality of human models, and of AIs meant to collaborate with humans.

\prg{Generalization to new environments.} Much work has been focused on creating agents that can perform well in a variety of environments, especially for the purposes of sim-to-real transfer \citep{zhao}. One common technique is \emph{domain randomization}, which involves defining a distribution over environments $\envdistr$ and training the agent to perform well in any randomly drawn environment~\citep{yu2017preparing, tan2018sim, jaderberg_human-level_2019}. We use this approach for our agent training in order to have it generalize to new environments at test time.

\section{Preliminaries}\label{preliminary}

\prg{Multi-agent MDPs.} A \emph{$n$-player MDP} $\mathcal{M}$ is defined by a tuple $\big\langle \mathcal{S}, \{\mathcal{A}_i\}_{1:n}, \mathcal{P}, \mathcal{T}, \gamma, \mathcal{R}, H \big\rangle$. $\mathcal{S}$ is a finite set of states; $\mathcal{A}_i$ is the finite set of actions available to agent $i$; $\mathcal{P}$ is the initial state distribution; $\mathcal{T}: \mathcal{S} \times \mathcal{A}_1 \times \dots \times \mathcal{A}_n \times \mathcal{S} \to [0, 1]$ is a transition function which specifies the distribution over the next state, given all agents' actions; $\gamma$ is the discount factor, and lastly $\mathcal{R}: \mathcal{S} \to \mathbb{R}$ is a real-valued reward function. We additionally denote the expected reward obtained by a policy tuple $\pi_1, \dots, \pi_n$ by $\rho(\pi_1, \dots, \pi_n) = \mathbb{E}_{\pi_1, \dots, \pi_n}[\sum^H_{t=0} r_t]$.

\prg{Embedding agents in multi-agent MDPs.} If we are given every agents' policy except for the $k$th policy $\{\pi_i : i \neq k\}$, then considering the multi-agent MDP $\mathcal{M}$ from the perspective of agent $k$, the other agents can be considered as ``part of the transition dynamics of the environment''. The problem of finding the optimal $\pi_k$ of this multi-agent MDP $\mathcal{M}$ can thus be reduced to finding the optimal $\pi$ for a single-agent MDP $\dot{\mathcal{M}}$ in which the transition dynamics compose the choices of actions by the other agents $\{\pi_i : i \neq k\}$, and the original transition dynamics $\mathcal{T}$ \citep{baskent_solving_2017}.

\prg{Human-AI collaboration through deep RL.}\label{sec:ha} In the simplest human-AI collaboration setting, we are given a two-player MDP $\mathcal{M}$ consisting of a human and an AI agent. If the human policy $\pi_H$ were known, we could simply embed the policy in the environment (as described above) and use RL to obtain an optimal collaborative policy. Given that sampling from the true human policy $\pi_H$ is expensive, training a deep RL policy with the human fully in-the-loop is unfeasible. The approach taken by previous work \citep{caroll2019utility, puig_watch-and-help_2020} is to leverage human data to \textit{learn} a human model $\hat{\pi}_H$, embed it into the environment, and then use deep RL to find an optimal policy for this induced single-agent MDP $\dot{\mathcal{M}}$.
If the human model is of sufficiently high quality (i.e. $\hat{\pi}_H$ is sufficiently close to $\pi_H$), the resulting agent policy will be able to collaborate with humans at test time.

\prg{Best Response (BR).} In a two-player collaborative game, the best response operator $\BR: \Pi \rightarrow \Pi$ takes in a fixed partner policy $\pi$ and returns an agent policy that maximizes the expected collaboration rewards $\rho$ when paired with $\pi$: $\BR(\pi) = \argmax_{\tilde{\pi} \in \Pi} \mathbb{E}[\rho(\tilde{\pi}, \pi)]$. Note that computing the optimal policy for the single-agent MDP $\dot{\mathcal{M}}$ above is equivalent to computing the best response to the human model $\pi_H$. In environments that are sufficiently complex, computing best responses exactly is unfeasible, leading to the usage of approximate methods: in our setting, deep RL can be considered our choice of \textit{approximate best response operator} $\BRa$ \citep{baskent_solving_2017}.

\section{Optimal Behavior Prior (\acr) for efficient human modelling}\label{sec:method}

\prg{Setup.} We want to obtain a human model of the highest possible quality given a limited set of human data that we have available. Let $\envdistr$ be a distribution over the possible environments that we are interested in deploying our system in. We have access to human-human gameplay data for $N$ environments from the distribution $\envdistr$. That is, we will have a dataset of a human collaborating with another human $\humandata^i$ with $i \in \{1,\dots,N\}$, for each environment $\sampleenv_i \sim \envdistr$. Note that when $\envdistr$ deterministically returns a single-environment, this setup is equivalent to a more traditional single-environment case.

\prg{Optimal-Behavior Prior ($\acr$).} We leverage the observation that -- from a Bayesian perspective -- \textit{the parameter initialization of a human model can be thought of as encoding a specific prior over the model space} \citep{maclaurin_early_2015, grant_recasting_2018}. The standard approach of randomly initializing parameters from high-capacity human models -- insofar as inputs will tend to output high-entropy distributions over actions -- is equivalent to a prior that humans will have highly random behavior. Our insight is that \textit{while human behavior is generally not optimal} (or one could use optimal agents as human models), in the vast majority of tasks, it is a fair assumption to expect \textit{humans to be better at the task than random behavior} (i.e. humans ``try to be optimal''). This leads us to expect that substituting the standard ``random behavior'' prior with an ``optimal behavior prior'' would facilitate the approximate Bayesian inference enacted by the model optimization \citep{maclaurin_early_2015}, potentially requiring significantly fewer iterations (and less data) to achieve the same level of human model quality. In practice, we show that even the simplest approach to encode such a prior -- using the model weights from a trained (near)-optimal agent as weight initialization for our human model -- works much better. 

\prg{Computing $\acr$.} Straightforwardly applying this idea requires the optimal behavior model (from which we intend to take the parameterization) and the human model to be of the same form. We consider neural network models, as they are plausibly expressive enough to represent both optimal and human behavior with the same model form. To obtain optimal behavior to use as prior for our multi-environment human model, we train a policy in self-play over the distribution of environments $\envdistr$, leading to a near-optimal agent over $\envdistr$, $\mlsp$, whose parameterization we can take and use to warm-start the training of our human model $\mlbcmlsp$ (see \Cref{fig:method-overview} and \Cref{alg:method}).

\prg{$\acr$ with freezing ($\acr+f$).} One can also view $\acr$ through the lens of representation learning~\citep{repr}: the representations learned by optimal agents will likely be more relevant to the task than randomly initialized ones. We investigate whether the intermediate representations learned by optimal agents are sufficient to model human behavior, by freezing the first $k$ network layers from the $\acr$ initialization and only fine-tuning the remaining network layers. From the Bayesian perspective, this can be seen as extra regularization, by encoding a Dirac prior on certain components of the model (i.e. complete certainty about the representations up until layer $k$) -- this assumption of absolute certainty about representations is thus implicitly equivalent to the assumption that the remaining free parameters still contain enough expressivity to represent human behavior.

\prg{Overview of assumptions.} To summarize, our main assumptions are: access to a distribution of environments of interest $\envdistr$, similar to ones we want to be able to collaborate with humans in at test time; access to human dataset from a limited number of diverse environments $D_{HH}^{1:N}$; being able to train (near-)optimal behavior over environments in $\envdistr$; being able to use the same model form for our human model as the (near-)optimal behavior parameterization. Finally, we are assuming that human behavior at the task at hand is closer to optimality than to chance.

Having obtained a general human model, we can then extend the deep RL method described in \Cref{sec:ha} to train collaborative AIs across the whole of $\envdistr$ through domain randomization. See \Cref{alg:method} for the full algorithm for training the OBP human model and the collaborative AI agent.

\begin{algorithm}
\caption{Behavior Cloning with OBP, and Human-Aware Training}
\begin{algorithmic}
\Require $\envdistr$, a distribution of environments; $\humandata^i$ for $i \in \{1,\dots,N\}$, human-human data for $N$ environments in $\envdistr$;

\State $\mlsp \leftarrow$ SelfPlay($\envdistr$)
\State $\mlbcmlsp \leftarrow$ $\mlsp$ \Comment{Initialize the behavior cloning policy with near-optimal behavior}
\State $\mlbcmlsp \leftarrow$ BehaviorCloning$\big(\mlbcmlsp$, $(\humandata^i)_{i=0}^N\big)$ \Comment{Fine-tune the initial policy on human data}
\State $\BRa(\mlbcmlsp) \leftarrow$ HumanAwareRLTraining$\big(\envdistr, \mlbcmlsp\big)$ \Comment{Train collaborative AI as in \citep{caroll2019utility}}

\end{algorithmic}
\label{alg:method}
\end{algorithm}

\section{Experimental setup}\label{sec:hypotheses}

\prg{Hypotheses.} In our experiments, we hope to investigate the following two hypotheses: 
 
\begin{description}[leftmargin=0.2in]
    \item \textbf{H1.} $\acr$ enables human models to better generalize to unseen environments, measured by reward similarity to H and validation loss. 
    \item \textbf{H2.} When trained with a human model using $\acr$ initialization, human-aware AI agents become better at collaborating with real humans on unseen environments, as measured by task reward. 
\end{description}

To validate \textbf{H1} and \textbf{H2}, we respectively consider a variety of possible training regimes for human models (\Cref{sec:humanmodels}), and collaborative agents (\Cref{sec:br-def}).

\subsection{Multi-environment human models}\label{sec:humanmodels}

We use high-capacity models -- neural networks -- to parameterize models of human behavior. Thanks to their expressivity, the same model form can also be used to parameterize (near-)optimal behavior. Below are the human model training regimes which we consider:

\prg{Vanilla BC ($\mlbc$).} We take a randomly initialized neural network, and perform behavior cloning on \textit{all} the collected human data $\humandata^{1:N}$ directly, i.e. supervised learning on the state-action pairs.

\prg{Multi-environment self-play ($\mlsp$).}
We train an agent on the distribution of environments $\envdistr$, incentivizing the policy to achieve high collaborative reward in self-play on all environments simultaneously. This $\mlsp$ agent is later used as proxy for optimal behavior. We include them in this section because $\mlsp$ is a baseline for human modeling: when no data is available to obtain a human model of good quality, using optimal behavior as a proxy for human behavior directly can sometimes be sufficient \citep{jaderberg_human-level_2019, OpenAIFive, AlphaStar, Strouse2021CollaboratingWH}. As a caveat, one can only expect this approach to work well in competitive environments, but has no guarantees in collaborative settings \citep{caroll2019utility}.

\prg{BC with optimal behavior prior ($\mlbcmlsp$).} To enable the multi-environment behavior cloning agents to more efficiently leverage human data, we encode the optimal behavior prior as described in \Cref{sec:method}, initializing the BC network with the $\mlsp$ agent's weight values.

\prg{BC with optimal behavior prior and freezing ($\mlbcmlspfreeze$).} We also consider an additional condition where all layers \textit{except} the last of the $\acr$ initialization are frozen, and only the last-layer weights are allowed to update using the human data. As described in \Cref{sec:method}, this allows us to apply extra regularization and understand whether the representations learned by the earlier layers of the parameterized $\acr$ prior need to be changed.
    
\subsection{Best Response (BR) agents}\label{sec:br-def}

In order to obtain collaborative AIs, we use deep RL to find an approximate best response $\BRa$ to a human model $\hat{H}$ as described in \Cref{preliminary}. However, in theory the $\acr$ method is agnostic to the approach used to compute the best response to the human model once it's obtained.

\prg{Multi-environment self-play: $\mlsp = \BRa(\mlsp)$.} As a baseline collaborative AI, we consider the multi-environment self-play agent from \Cref{sec:humanmodels}. By definition, such agent is a best response to itself: $\BRa(\mlsp) = \mlsp$. We will thus refer to $\BRa(\mlsp)$ as $\mlsp$ -- this is the agent we use in practice.

\prg{Multi-environment human-aware agents: $\BRa(\mlbc)$, $\BRa(\mlbcmlsp)$, and  $\BRa(\mlbcmlspfreeze)$.} These collaborative AI agents are instead explicitly trained to complement human models based on human data, by training an approximate best response on environments from the multi-environment distribution $\envdistr$ via deep RL (as described in \Cref{sec:ha}).
We do this for all human-data-based human models defined in \Cref{sec:humanmodels}, obtaining $\BRa(\mlbc)$, $\BRa(\mlbcmlsp)$, and $\BRa(\mlbcmlspfreeze)$.


\subsection{Simulated user study}\label{exp_setup}

\begin{figure*} [t]
\vskip -0.6em
    \centering
     \begin{subfigure}[b]{0.193\textwidth}
         \centering
         \includegraphics[width=\textwidth]{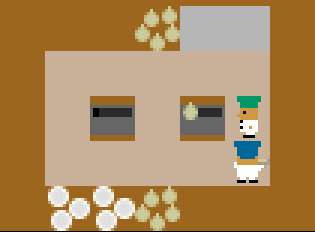}
     \end{subfigure}
     \hfill
     \begin{subfigure}[b]{0.193\textwidth}
         \centering
         \includegraphics[width=\textwidth]{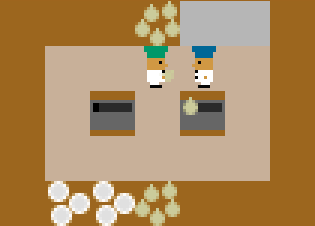}
     \end{subfigure}
     \hfill
     \begin{subfigure}[b]{0.193\textwidth}
         \centering
         \includegraphics[width=\textwidth]{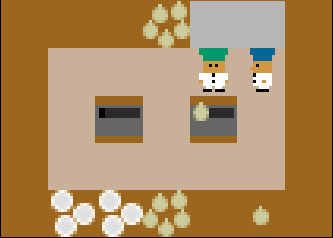}
     \end{subfigure}
     \hfill
    \begin{subfigure}[b]{0.193\textwidth}
         \centering
         \includegraphics[width=\textwidth]{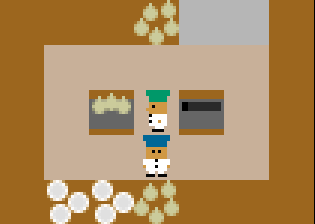}
     \end{subfigure}
     \vskip -0.2em
     \caption{\prg{Example game-play in \textit{Overcooked-AI}.} To successfully deliver a dish, the two chefs need to pick up and place 3 onions in a pot, kick off the cooking process, pick up the resulting soup with a dish, and drop it off at a serving station. The goal is to maximize soup deliveries within 400 timesteps.}
     \vskip -0.9em
\label{fig:overcooked-ai}
\end{figure*}

\prg{Framework.} For all our experiments, we use the Overcooked-AI environment framework from \cite{caroll2019utility}, which was designed as a test-bed for human-AI collaboration performance, and has been used in other various works \citep{nalepka_interaction_2021, fontaine_importance_2021, Ribeiro2022AssistingUT}. In Overcooked-AI environments, two cooks in a kitchen gridworld collaborate to cook as many soups as possible within a time limit. See \Cref{fig:overcooked-ai} for an example.

\prg{Why a simulated user study?} 
We evaluate our hypothesis that \acr\ leads to better generalization by generating data from a ``ground truth human proxy'', and using that data for training and testing. 
We do this because having access to the ``ground truth human'' enables us to run more extensive analysis that wouldn't be possible with real people: for example, we can see how good our human model is relative to good it could possibly be (e.g. maybe the ground truth is so noisy that anything beyond a certain level of predictive accuracy is impossible), in addition to testing how much better our model is relative to other baselines. Further, it lets us analyze the importance of the optimal-behavior prior in isolation, without confounding factors encountered when interacting with real users, such as their level of adaptation to agent. 

\prg{Ground truth human proxy (H).} On a high-level, our ground truth human proxy model has different parameters which are fit to real human data (or manually selected to induce human-like behavior). Its performance is no-where near optimal (compare H to $\mlsp$ reward in \Cref{fig:val-comp-bc}), as could be expected for actual human performance. This suggests that our ground truth human proxy won't benefit by using $\acr$ relative to real humans. While in \Cref{sec:exp} we only show results for one ground truth human, in \Cref{sup:cem-info} we describe other two mostly orthogonal ways of obtaining simulated humans which in preliminary experiments gave similar results to the ones in \Cref{sec:exp}, giving more credence to the hypothesis that this approach could also generalize to working with real humans.

\begin{figure*} [t]
    \vskip -0.8em
    \centering
     \begin{subfigure}[b]{0.193\textwidth}
         \centering
         \includegraphics[width=\textwidth]{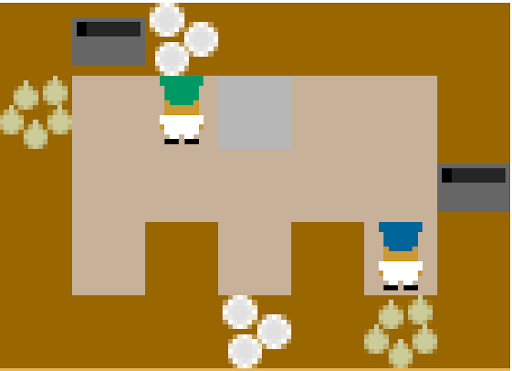}
         \caption{$\testlayout_0$}
         \label{fig:e0}
     \end{subfigure}
     \hfill
     \begin{subfigure}[b]{0.193\textwidth}
         \centering
         \includegraphics[width=\textwidth]{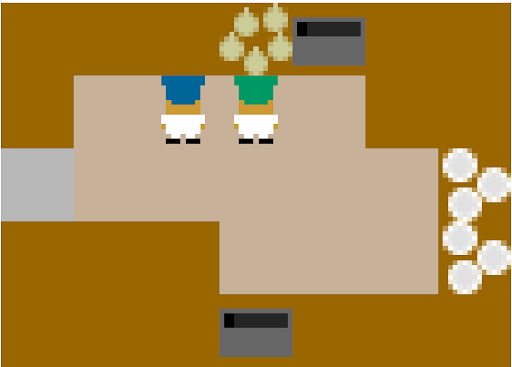}
         \caption{$\testlayout_1$}
         \label{fig:e1}
     \end{subfigure}
     \hfill
     \begin{subfigure}[b]{0.193\textwidth}
         \centering
         \includegraphics[width=\textwidth]{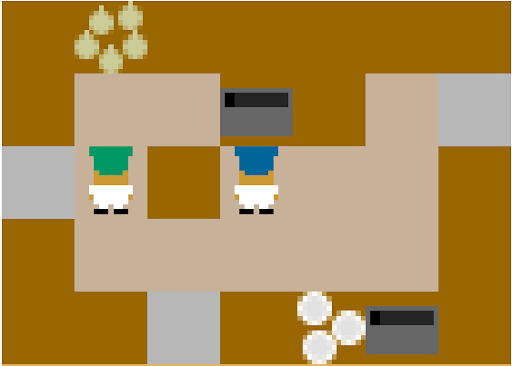}
         \caption{$\testlayout_2$}
         \label{fig:e2}
     \end{subfigure}
     \hfill
    \begin{subfigure}[b]{0.193\textwidth}
         \centering
         \includegraphics[width=\textwidth]{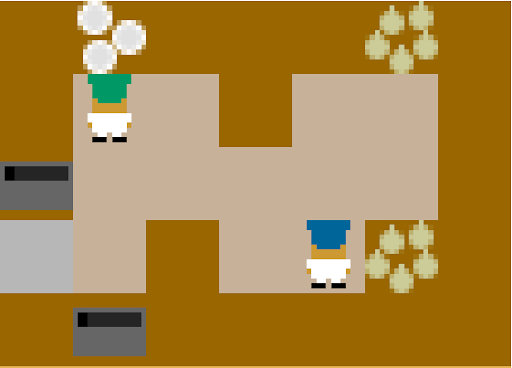}
         \caption{$\testlayout_3$}
         \label{fig:e3}
     \end{subfigure}
     \hfill
     \begin{subfigure}[b]{0.193\textwidth}
         \centering
         \includegraphics[width=\textwidth]{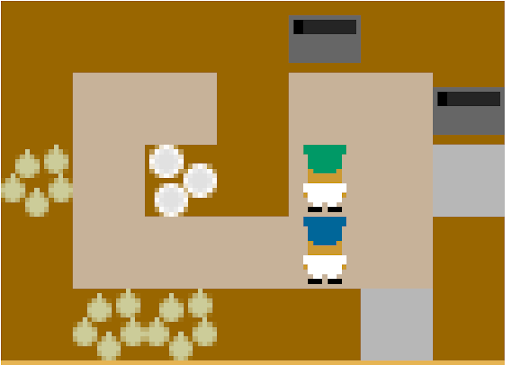}
         \caption{$\testlayout_4$}
         \label{fig:e4}
     \end{subfigure}
     \vskip -0.2em
     \caption{\textbf{Our evaluation layouts.} The evaluation layouts were selected to cover the entire spectrum of layout difficulties (where $\testlayout_0$ is easiest and $\testlayout_4$ is hardest). In the easier layouts (left), two agents can perform the required tasks independently in different areas of the space without needing to interact much. In the harder layouts (right), the two agents need to coordinate with each other to get through narrow pathways without colliding.}
     \label{fig:eval-layouts}
     \vskip -1em
\end{figure*}

\prg{Distribution of layouts.} All $\sim 10^{13}$ layouts in the training environment distribution $\envdistr$ are $7 \times 5$ in size, but differ in location of walls, pots, and dispensers (see \Cref{sup:multilayoutok}). 
For the evaluation layouts, we selected 5 layouts from $\envdistr$: $\testlayout_0, \testlayout_1, \testlayout_2, \testlayout_3, \testlayout_4$ in the neighborhood of 100th, 75th, 50th, 25th and 0th percentile of maximum achievable self-play rewards (see \Cref{fig:eval-layouts}). Selecting layouts based on achievable scores -- instead of random sampling -- gives us insights into the performance of the human models and best responses across the spectrum of layout difficulties. Evaluating on a small set of test environments also allows us to compare multi-env approaches with single-env ones (\Cref{sec:sl-comp}).

\prg{Simulated human data.} By pairing the simulated human $H$ with itself in $N=40$ layouts for 1 game per layout, we obtain a multi-layout dataset $D_{HH}^{1:N}$. This dataset is then equally split into a training set $D_{HH}^{\text{train}}$ and a test set $D_{HH}^{\text{test}}$.

\prg{Human model and $\BRa$ model.}\label{sec:nn-info} All human models and all collaborative agents share the same neural network architecture: the models take in the full Overcooked-AI game state and output a probability distribution over 6 possible actions: 4 directions (e.g. north), wait, and interact. Taking in a full game-state featurization (see \Cref{sup:nn-info}) is an important difference from prior work on human modeling in Overcooked-AI \citep{caroll2019utility, paulknott}, which uses a simplified handcrafted state featurization to make behavior cloning be sufficiently data-efficient for any meaningful learning too occur. Given our multi-env setting, this is a necessary choice: developing a good hardcoded featurization which would be sufficient for a whole distribution of layouts would be very challenging in this domain.
All BC variants perform behavior cloning on $D_{HH}^{\text{train}}$ by minimizing the cross-entropy loss, and early-stop when the cross-entropy loss on $D_{HH}^{\text{test}}$ starts to increase. Additional information for BC training conditions is located in \Cref{sup:bc-conditions}.
For collaborative AIs, we perform $\BRa$ training with PPO~\citep{schulman_proximal_2017}. In the case of multi-layout $\BRa$ agents, every on-policy PPO rollout is sampled from a different environment $\sampleenv \sim \envdistr$ (see \Cref{sup:ha-conditions}).


\section{Experimental results}\label{sec:exp}

\subsection{Human modeling results}\label{sub:hm_results}

To test the effectiveness of our method, we tested imitation learning using OBP or not, whether we froze the representation layers ($+ f$), and the usage of multi-layout human data (BC or SP).
We measure the rewards obtained by pairing the human model with a copy of itself on the evaluation layouts; we also measure the validation loss on our human data from the evaluation layouts, $D_{HH}^{\text{test}}$. See \Cref{fig:eval-layouts}.

\prg{$\mlsp$ is a bad human model under all metrics considered.} As a first observation, we confirm that self-play does very poorly according to our ``realism metrics'' for human models: both validation loss and rewards obtained by $\mlsp$ agents are drastically higher than those of the ground truth human model proxy. Qualitatively, $\mlsp$ can also immediately be identified as non-human (and specifically \textit{superhuman}): real humans (and the ground truth human proxy) chooses ``wait'' actions around 55\% of timesteps (due to the game's high clock-speed), while $\mlsp$ never does (as a near-optimal agent).

\prg{$\mlbc$ without $\acr$ gets zero reward.} Since the reward is sparse, zero reward means that the agents never completed an entire delivery (while we do qualitatively observe them perform initial tasks, they eventually get stuck). Poor human modeling performance relative to prior work can be explained by the greater difficulty of modelling human behavior across many different layouts, plus training using the full state representation (of dimensions $26 \times 7 \times 5$) rather than a simplified $66$-dimensional manual featurization (as explained in \Cref{sec:nn-info}).

\begin{figure*}[t]
\vskip -1.2em
\centering
\begin{subfigure}[b]{0.49\textwidth}
   \includegraphics[width=\linewidth]{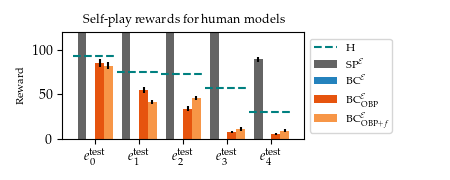}
\end{subfigure}
\begin{subfigure}[b]{0.49\textwidth}
   \includegraphics[width=\linewidth]{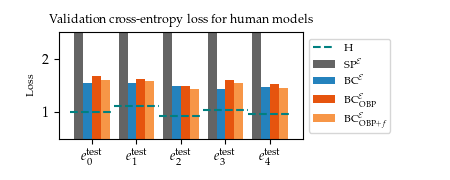}
\end{subfigure}
\vskip -1.1em
\caption{\textbf{(Left) Rewards.} Note that $\mlsp$ achieves super-human performance and thus unlikely to be human-like, whereas $\mlbc$ \textit{obtains 0 reward on all evaluation layouts}. \textbf{(Right) Validation loss.} We see that simple $\acr$ tends to do worse than $\acr + f$ and simple $\mlbc$. For an in-depth analysis of why simple $\mlbc$ performs so well, see \Cref{sup:human-modeling}. For both plots, the dotted line is the ``gold standard'', the ground truth human reward and loss on the data H generated itself.}
\label{fig:val-comp-bc}
\vskip -0.9em
\end{figure*}

\prg{$\acr$ significantly improves the reward, but not the loss.} Quite surprisingly, although using $\acr$ leads the human model to achieve significantly higher reward (which is much closer to the ground truth human's reward, and thus ``reward-realistic''), the cross-entropy losses are almost identical across all BC variants. By performing a more in-depth analysis, we found that BC without $\acr$ arrives at low cross-entropy solutions mostly by assigning excessive probability to the human's probability of waiting (as most human actions are ``wait''). For a detailed description of this phenomenon, see \Cref{sup:human-modeling}.
Qualitatively, human models using $\acr$ appear to be the most human-like, as they are actually able to complete the task, while simple $\mlbc$ is not able to complete any deliveries when paired with itself, and as mentioned above $\mlsp$ appears clearly superhuman.

\prg{Representation freezing.} Overall, freezing seems to be somewhat beneficial: it seems to increase rewards especially for harder layouts, and slightly lowers validation loss across the board when used with $\acr$, leading to validation losses which are very similar (or marginally better) than simple BC (for a more in-depth analysis see \Cref{sup:human-modeling}). This goes to show that the extra regularization applied by freezing further improves the data-efficiency of the human model training -- albeit slightly -- and that thus optimal agent and human representations can be shared quite easily.

Overall, the qualitative results support \textbf{H1}, and so does the reward metric, while the validation loss results are inconclusive. We provide a more detailed discussion about this, and human model ``quality metrics'' more generally in \Cref{sec:discussion}.

\subsection{Collaborative AI results}

In \Cref{sub:hm_results} we showed that using \acr\ can improve human model quality -- but does using such improved models also improve collaborative AIs who best respond to them?

\begin{figure*}[t]
\vskip -1.5em
\centering
\begin{subfigure}[b]{0.49\textwidth}
   \includegraphics[width=\linewidth]{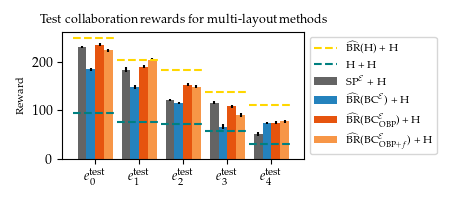}
\end{subfigure}
\begin{subfigure}[b]{0.49\textwidth}
   \includegraphics[width=\linewidth]{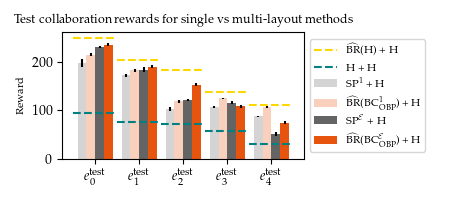}
\end{subfigure}
\vskip -0.8em
\caption{\textbf{(Left) Multi-layout conditions.} We display the test collaboration rewards for best responses (BR) to different types of human proxies, with standard errors computed over 40 rollouts. The teal horizontal line shows the average Human-Human collaboration rewards, and the gold standard of performance is obtained by an agent trained with H directly on the evaluation layout. We can see that in general, human-aware methods achieve higher collaboration reward when paired with H.
\textbf{(Right) Single-layout comparison.} We see that $\BRa(\mlbcmlsp)$ achieves comparable or better performance to $\BRa(\slbcslsp)$ on the majority of the evaluation layouts, showing that deep RL can converge to better optima when performing domain randomization.
}
\label{fig:test-rew-br}
\vskip -1.1em
\end{figure*}

For deep RL training, we varied the human models to which we are computing a $\BRa$ to ($\mlsp$, $\mlbc$, $\mlbcmlsp$, and $\mlbcmlspfreeze$). This respectively leads to the following conditions: $\mlsp, \BRa(\mlbc)$, $\BRa(\mlbcmlsp)$, and $\BRa(\mlbcmlspfreeze)$.
We measure collaboration rewards by pairing trained $\BRa$ agents with the ground truth proxy human $H$ on our 5 evaluation layouts. See \Cref{fig:test-rew-br}.


\prg{Using $\acr$ leads to best collaboration reward.} Using $\acr$-based human models enables the best responses $\BRa(\mlbcmlsp)$ and $\BRa(\mlbcmlspfreeze)$ to achieve the highest human-collaboration rewards on almost all layouts (although sometimes by slim margins); this supports the claim that \textit{when trying to obtain collaborative AI agents, it's often advantageous to use the improved human models obtained by} $\acr$, $\mlbcmlsp$ \textit{and} $\mlbcmlspfreeze$, \textit{instead of assuming human optimality directly} ($\mlsp$). Qualitatively we find that $\mlsp$ agents often force their human collaborators to change their trajectories, whereas the $\BRa(\mlbcmlsp)$ and $\BRa(\mlbcmlspfreeze)$ agents stay out of the human's way, leading to a smoother collaboration experience.

\prg{SP does surprisingly well relative to $\BRa(\mlbc)$.} While this is seemingly in contradiction with prior work, we found this to be because $\mlbc$ was learned without a handcrafted featurization of the state (as mentioned in \Cref{sec:nn-info}), leading the resulting human model $\mlbc$ to be ``low quality'' (at least in terms of reward). Overall, one insight from this is that \textit{unless you are able to obtain a sufficiently high-quality human model, assuming human optimality (i.e. using self-play models as human models, and thus also as best responses) seems to work well in practice in Overcooked-AI.} This effect might be particularly marked in our distribution of environments $\envdistr$ given that the main source of human suboptimality for such relatively simple tasks is their speed (as mentioned before, more than half of all actions tend to be ``wait''): on a strategic level, real humans and self-play agents don't seem to differ much in their gameplay (see \Cref{sup:human-modeling}).

Our simulated user study mostly supports \textbf{H2}, giving hope that it would also hold with real humans.

\prg{Corresponding single-layout agents.}\label{sec:sl-comp} To further investigate the effectiveness of our approach, we introduce single-layout counterparts to the multi-layout agents defined in \Cref{sec:br-def}: $\slsp$ and $\slbcslsp$. Unlike the multi-layout agents, all single-layout agents have direct access to both the evaluation layout $e_i$ and human data on the evaluation layout $D_{HH}^{\testlayout_i}$ during training, and thus all the human-aware training happens on $\testlayout_i$ with BC partner who had access to $D_{HH}^{\testlayout_i}$. See \Cref{fig:test-rew-br} for results.

\prg{Multi-layout agents can do better than the single-layout ones.} In most cases, the zero-shot performance of $\BRa(\mlbcmlsp)$ on evaluation layouts is comparable to its single-layout counterpart $\BRa(\slbcslsp)$ (which had access to the layout for training!) -- and on easier layouts it can even be significantly better. This speaks to the approximate nature of our $\BRa$ operator, deep RL. Perhaps surprisingly, multi-layout training seems to improve the quality of the optimum reached by deep RL, potentially thanks to the increased robustness and improved representations required to do well on an entire distribution of layouts relative to a simple one. This seems to suggest that \textit{making the RL problem harder through multi-environment training can also play a role in improving downstream human-AI collaboration, when policies are obtained through deep RL}.

\section{Discussion}\label{sec:discussion}
\prg{Metrics for evaluating human models.} One takeaway from our experiments is that it's not obvious how one should judge the quality of human models. Human models can have the exact same loss (for human data prediction) but lead to radically different qualitative behavior. Similarity in reward to humans seems like a potentially better metric, but it's also not bulletproof: in the most extreme case, memorizing the training data would lead to the exact same reward as the demonstrations. 

\prg{Limitations and future work.} Firstly, while we performed our experiments with simulated ground truth proxy humans, more work is required to verify that these results will hold with real humans. In addition, a clear avenue of future work would be that of exploring other behavior priors which might be closer to human than optimal behavior, including but not limited to max-entropy agents. Another interesting avenue of future work regards the warm-starting of the training of optimal agents: in complex domains, it's common to bootstrap RL with a imitation learning policy to speed up training \citep{OpenAIFive, AlphaStar}. Inspired by this, we speculate whether it may be advantageous to combine these approaches, first training a simple BC model (which is very fast) which can be used to bootstrap RL training (as it helps address exploration problems), and then -- assuming that the information in the human data would be lost to catastrophic forgetting -- one could using the resulting RL policy to boostrap the training of a better human model (via OBP).

\prg{Summary.} In our work, we try to create generalizable human models for the purposes of improving human modeling and human-AI collaboration. While straightforward multi-environment human modeling struggles to learn anything meaningful, we find that good quality multi-environment human modeling is rendered possible by leveraging the knowledge that humans will be better than random with an optimal behavior prior. We then show that this type of modeling approach is also advantageous for downstream collaborative reward, performing better than modeling the human as optimal.

\begin{ack}

We'd like to thank members of the InterACT lab for helpful discussion. This work was supported by NSF CAREER, NSF NRI, ONR YIP, and the NSF Fellowship.

\end{ack}

\bibliography{main}
\bibliographystyle{plainnat}

\newpage
\appendix

\section{Additional information about multi-Layout \textit{Overcooked-AI}}\label{sup:multilayoutok}

\begin{figure}[b]
\centering
\begin{tabular}{ccccccc}
{\includegraphics[width = 0.11\textwidth]{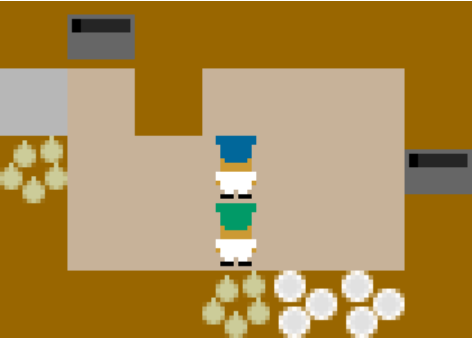}} &
{\includegraphics[width = 0.11\textwidth]{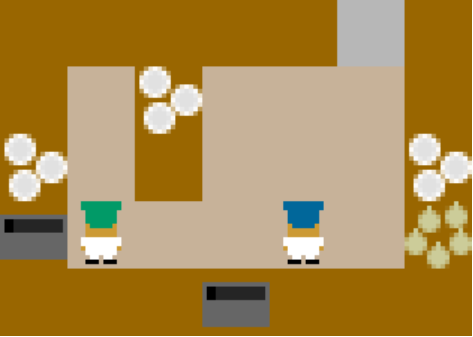}} &
{\includegraphics[width = 0.11\textwidth]{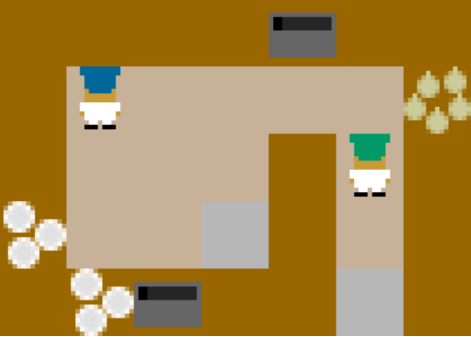}} &
{\includegraphics[width = 0.11\textwidth]{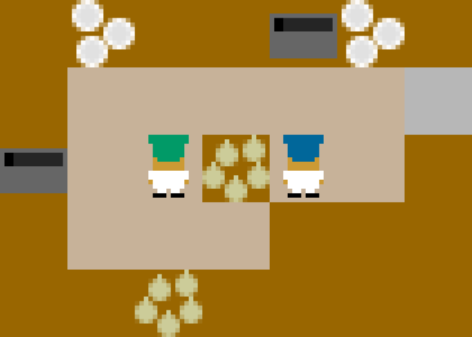}} &
{\includegraphics[width = 0.11\textwidth]{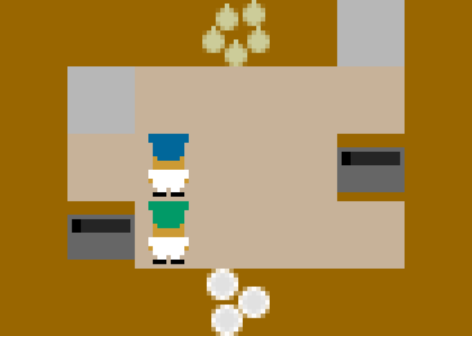}} &
{\includegraphics[width = 0.11\textwidth]{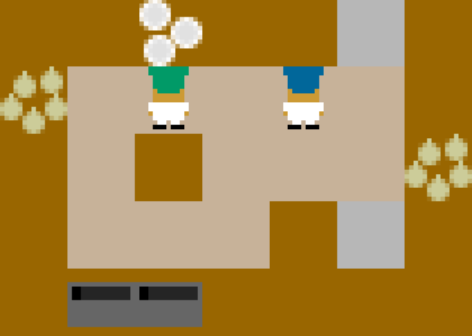}} &
{\includegraphics[width = 0.11\textwidth]{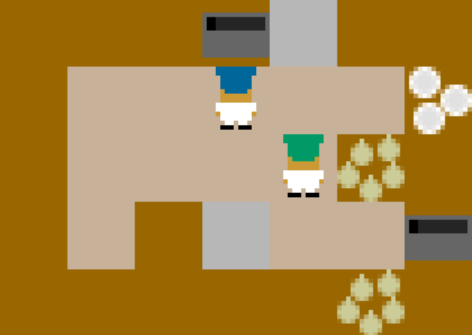}} \\
{\includegraphics[width = 0.11\textwidth]{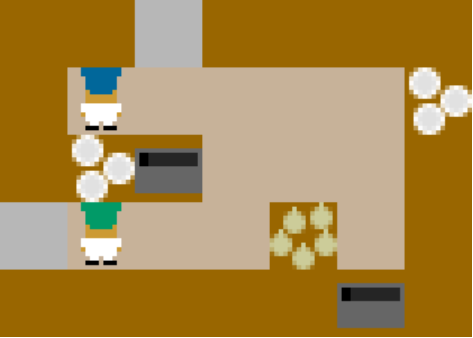}} &
{\includegraphics[width = 0.11\textwidth]{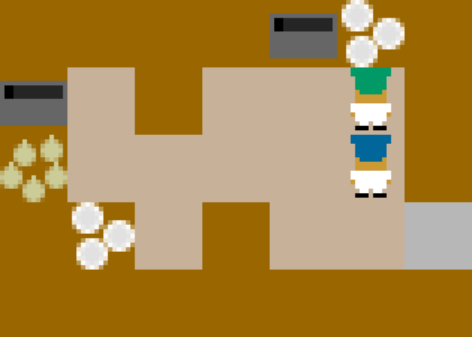}} &
{\includegraphics[width = 0.11\textwidth]{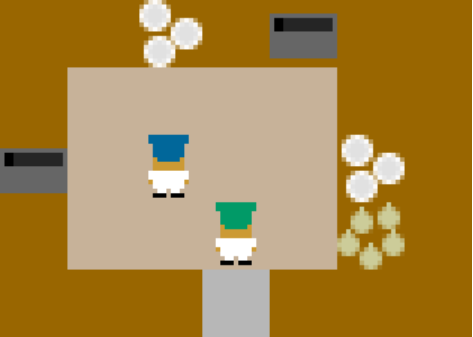}} &
{\includegraphics[width = 0.11\textwidth]{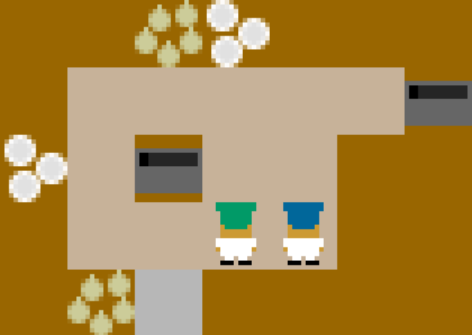}} &
{\includegraphics[width = 0.11\textwidth]{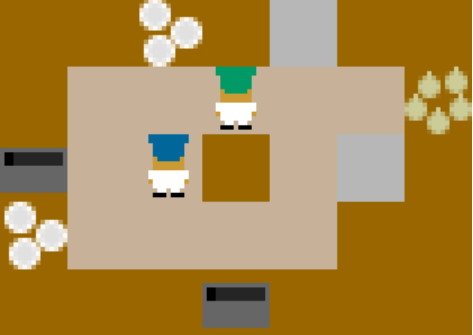}} &
{\includegraphics[width = 0.11\textwidth]{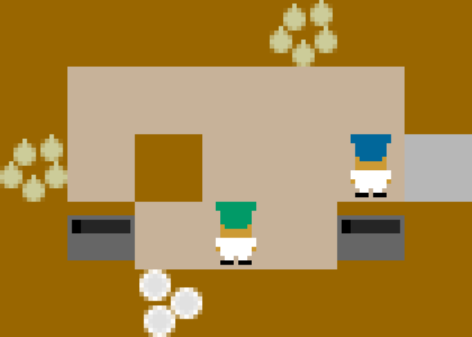}} &
{\includegraphics[width = 0.11\textwidth]{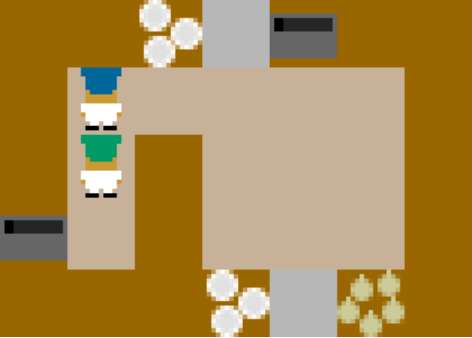}} \\
{\includegraphics[width = 0.11\textwidth]{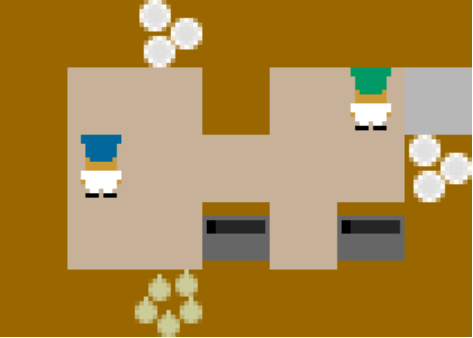}} &
{\includegraphics[width = 0.11\textwidth]{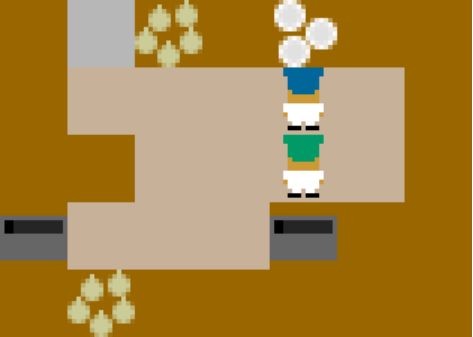}} &
{\includegraphics[width = 0.11\textwidth]{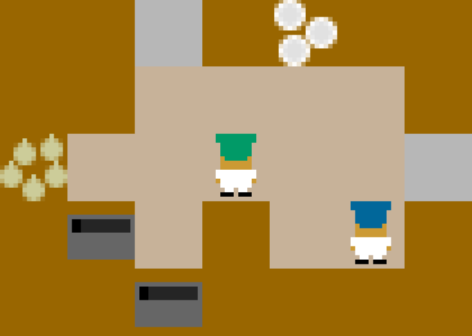}} &
{\includegraphics[width = 0.11\textwidth]{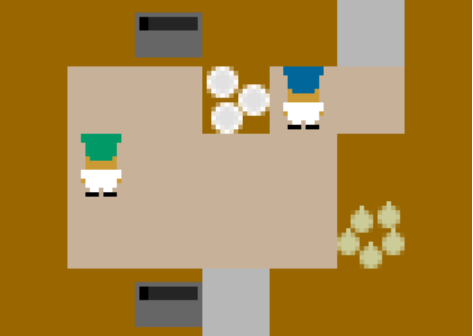}} &
{\includegraphics[width = 0.11\textwidth]{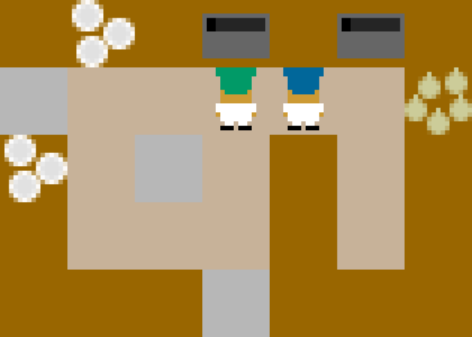}} &
{\includegraphics[width = 0.11\textwidth]{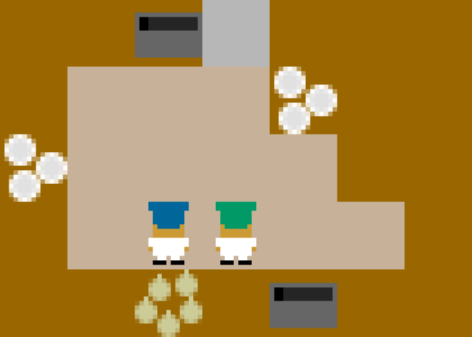}} &
{\includegraphics[width = 0.11\textwidth]{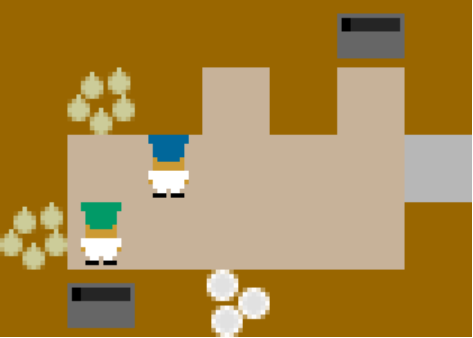}} \\
\end{tabular}
\caption{\textbf{Samples from the distribution of training environments $\envdistr$.} Light brown squares are empty spaces that the cooks can occupy. Dark brown squares are counters and can be occupied by an onion dispenser (yellow shapes), a dish dispenser (white circles), pot (a dark grey rectangle with black top), and serving station (a light grey square).}
\label{fig:21sample}
\end{figure}

The multi-layout distribution contains roughly $10^{13}$ $7 \times 5$ layouts of approximately 75\% empty space. Roughly 40\% of all counters are occupied by an interactive object (onion dispenser, dish dispenser, pot, or serving station). A cook can move into any adjacent empty square that is not occupied by its collaborator, and they can also interact with any object by facing the object and choosing the \textit{interact} action. See \apfigref{fig:21sample} for additional details.

\section{Model form details}\label{sup:nn-info}

\subsection{Human models and best response agents}

All human models and best response agents share the same neural network structure. The network takes a lossless state encoding with shape $26 \times 7 \times 5$, initially processed by 3 convolutional layers (with 25 kernels per layer, each of size $5 \times 5$, $3 \times 3$, and $3 \times 3$ respectively) followed by 3 layers of fully connected layers (of hidden size 64), and output a probability distribution over the legal actions: north, south, east, west, wait, and interact (a $6$-dimensional vector). While it might improve performance in practice, we don't use recurrence over state history to reduce the computational burden of training.

\subsection{Simulated humans}\label{sec:sim-human-info}

We run our experiments with simulated humans. At every timestep, such models re-plan an action sequence to complete the next high level task (selected greedily), and take the first action in that plan. They have three tunable parameters: \textbf{1) probability of waiting} ($prob\_{wait}$): This is the probability of not moving (taking the "wait" action) at each time-step. \textbf{2) low-level Boltzmann irrationality} ($ll_{temp}$) which injects noise in the low-level action selection process while carrying out the picked goal, where the noise is proportional to the suboptimality of that action (i.e. causing the model to sometimes pick a "left" or "wait" action instead of "right"). \textbf{3)} \textbf{high-level Boltzmann irrationality} ($hl_{temp}$) which injects noise in the goal-picking process according, once again with noise proportional to the goal cost (i.e. leading the agent to sometimes get onions from a more distant onion dispenser). 

In preliminary investigations, we tested different types of simulated humans, which differed in what aspect of the human's data they were fit to: \textbf{1)} The \textbf{waiting-matching} $\text{H}_0$ only had it's probability of weighting (which is a parameter) fit to humans' proportion of waiting time; the low and high level noise are instead set to 0. \textbf{2)} The \textbf{action-matching} $\text{H}_1$ additionally fit to capture real-humans' suboptimality on the motion and goal-selection level (i.e. minimizing cross-entropy loss with respect to real human actions in the data). \textbf{3)} The \textbf{reward-matching} $\text{H}_2$ mainly focuses on achieving human-level reward -- only minimizing the reward difference between real humans and that obtained by the model. In preliminary experiments with all 3 models, we found our method to perform similarly relative to baselines. Throughout the main text, we let $H = H_0$.

\section{Training details}

\subsection{Simulated humans}\label{sup:cem-info}

As described above, in an attempt to make simulated human models be as human-like as possible, we fit their parameters from human data by running cross-entropy methods (CEM) ~\citep{Rubinstein1999TheCM} on the human-data collected in \cite{paulknott}. \aptabref{tab:hsim-all-conditions} shows the fitted parameters.

In preliminary experiments with all 3 models, we found our method to perform similarly relative to baselines. Throughout the main text, we let $H = H_0$.

\begin{table}[h]
    \centering
    \begin{tabular}{ |p{3.01cm}||p{0.85cm}|p{0.73cm}|p{1.5cm}|p{2.2cm}|p{2.3cm}|}
    \hline
    H name & $hl_{temp}$ & $ll_{temp}$ &  $prob\_{wait}$ & Average Cross-Entropy loss & Average reward fraction\\
    \hline
    H$_{0}$ (H in main results) & \includegraphics[width=6pt]{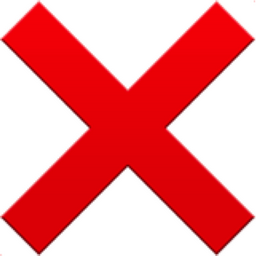} & \includegraphics[width=6pt]{images/emoji/cross-mark.png} & 0.5 & 1.62 & 0.631\\
    \hline
    H$_{1}$ & 0.072 & 0.286 & 0.45 & 1.81 & 0.318\\
    \hline
    H$_{2}$ & 0.070 & 0.249 & 0.04 & 3.19 & 0.805\\
    \hline
    \end{tabular}
    \vskip 0.6em
    \caption{\textbf{Three types of simulated humans} $\text{H}_0$ achieves the lowest cross-entropy loss while obtaining a decent level of reward-matching. $\text{H}_1$ achieves similar cross-entropy loss while achieving lower reward. $\text{H}_2$ achieves the most human-like reward at the expense of a much higher cross-entropy loss.}
 \label{tab:hsim-all-conditions}
\end{table}

\subsection{Human models and best response agents}\label{sup:training-info}

\prg{Self-play training.}\label{sup:sp-conditions} Multi-layout self-play is performed with a batch-size of 200k. Each batch is composed of 500 episodes of length 400, collected on 250 different layouts sampled from $\envdistr$. We train for 3000 epochs (8 SGD gradient steps with mini-batch of 160k) with PPO. Single-layout self-play training is also performed with a batch-size of 200k, but less training epochs are required: we use 500. \aptabref{tab:spcondition} provides a summary.

\begin{table}[h]
    \centering
    \begin{tabular}{ |p{1.49cm}||p{1.88cm}|p{1.65cm}|p{1.41cm}|p{1.0cm}|p{2.3cm}|p{1.1cm}|  }
    \hline
    SP method  & Training distribution & Number of layout seen & Batch size & Epochs &  Entropy coefficient schedule & Entropy horizon\\
    \hline
    $\mlsp$ & $\multi$ & $1.5 \times 10^6$ & 200k & 3000 & 0.2 $\rightarrow$ 0.0005 & 2000 ep\\
    \hline
    $\slsp$ & $\testlayout_i$ & 1 & 200k & 500 & 0.2 $\rightarrow$ 0.005 & 150 ep\\
    \hline
    \end{tabular}
    \vskip 0.6em
    \caption{Training conditions for SP. We find entropy schedule to make the biggest difference in the multi-layout setting relative to the single-layout one, while other PPO hyper-parameters are kept the same. $\mlsp$ achieves the same level of reward as $\slsp$ on all evaluation layouts.}
 \label{tab:spcondition}
\end{table}

\prg{Simulated data.} We use our "ground truth" human proxy to create a dataset as if we had run an actual user study with 20 pairs of participants. For each pair of participants, we would want to have them collaborate on 7 layouts for one episode -- where 5 layouts will be the test ones and the remaining two will be sampled from $\envdistr$. The intention behind this is that it would enable us to train human models based on data from the $\envdistr$, and also have ground truth data from the evaluation layouts. 

We simulate performing an entire user study as follows: we pair $H$ with itself on 7 layouts ($\testlayout_0$, $\testlayout_1$, $\testlayout_2$, $\testlayout_3$, $\testlayout_4$, and two sampled layouts $e^{\text{sampled}} \sim \multi$), and for each we collect an episode of length 400. We then repeat this procedure 20 times with the same pair of $H$ agents (but sampling different $e^{\text{sampled}}$ layouts from $\envdistr$ every time).

We then aggregate the data on the $N = 40$ sampled layouts $e^{\text{sampled}}_{0:39}$ (the probability of resampling the same layout is negligible) to form the multi-layout human data set $D^{1:N}_{HH}$, corresponding to the notation introduced in Sec 4 of the main text.

\prg{Behavior cloning.}\label{sup:bc-conditions} Multi-layout behavior cloning is performed on  the human-human play described above, which amounts to 32k game timesteps across the distribution of layouts. The single-layout BC counterparts in our experiment use 16k timesteps. For reference, in previous work~\citep{paulknott}, roughly 24k timesteps were used to train human models on a single layout. \aptabref{tab:bc-summary} provides a summary.

\begin{table}[h]
    \centering
    \begin{tabular}{ |p{1.6cm}||p{1.63cm}|p{1.2cm}|p{1.66cm}|p{2.55cm}|p{2.53cm}| }
    \hline
    BC method & Behavioral prior & Freezing & Human data & Final training loss & Validation loss \\
    \hline
    $\mlbc$ & \includegraphics[width=6pt]{images/emoji/cross-mark.png} & \includegraphics[width=6pt]{images/emoji/cross-mark.png} & 32k on $\multi$ & 1.34 & 1.60 (avg for $\testlayout_{1:5}$)\\
    $\mlbcmlsp$ & \includegraphics[width=6pt]{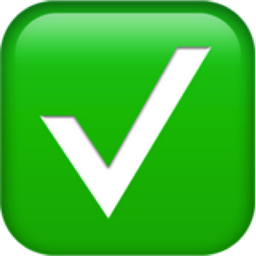} ($\mlsp$) & \includegraphics[width=6pt]{images/emoji/cross-mark.png} & 32k on $\multi$ & 1.09 & 1.70 (avg for $\testlayout_{1:5}$)\\
    $\mlbcmlspfreeze$ & \includegraphics[width=6pt]{images/emoji/white-heavy-check-mark.png} ($\mlsp$) & \includegraphics[width=6pt]{images/emoji/white-heavy-check-mark.png} & 32k on $\multi$ & 1.23 & 1.61 (avg for $\testlayout_{1:5}$)\\
    \hline
    $\slbc$ & \includegraphics[width=6pt]{images/emoji/cross-mark.png} & \includegraphics[width=6pt]{images/emoji/cross-mark.png} & 16k on $\testlayout_i$ & 1.02 (avg for $\testlayout_{1:5}$) & 1.10 (avg for $\testlayout_{1:5}$)\\
    $\slbcslsp$ & \includegraphics[width=6pt]{images/emoji/white-heavy-check-mark.png} ($\slsp$) & \includegraphics[width=6pt]{images/emoji/cross-mark.png} & 16k on $\testlayout_i$ & 0.99 (avg for $\testlayout_{1:5}$) & 1.12 (avg for $\testlayout_{1:5}$)\\
    \hline
    \end{tabular}
    \vskip 0.6em
    \caption{Training conditions for all BC methods, split into the multi-layout and single-layout groups. OBP allows the agent to achieve lower training loss. Single-layout variants are capable of achieving lower loss, likely thanks to memorization with a over-parameterized BC network.}
 \label{tab:bc-summary}
\end{table}

\prg{Best Response training.}\label{sup:ha-conditions} The rollouts of partnered-play (where its partner is fixed as the BC agent) episodes with length 400 are collected simultaneously on 500 freshly sampled layouts per iteration to form a batch-size of 200k, for 500 iterations. \aptabref{tab:br-summary} provides a summary.

\begin{table}[h]
    \centering
    \begin{tabular}{ |p{2.2cm}||p{1.7cm}|p{1.7cm}|p{1.4cm}|p{2.3cm}|p{1.3cm}|  }
    \hline
    $\BR$ method & Behavioral prior & Training distribution & Epochs & Entropy coefficient schedule & Entropy horizon \\
    \hline
    $\BRa(\mlbc)$, $\BRa(\mlbcmlsp)$, $\BRa(\mlbcmlspfreeze)$ & \includegraphics[width=6pt]{images/emoji/white-heavy-check-mark.png} ($\mlsp$) & $\multi$ &  500 &
    0.1 $\rightarrow$ 0.0005 & 250 ep\\
    \hline
    $\BRa(\slbc)$, $\BRa(\slbcslsp)$ & \includegraphics[width=6pt]{images/emoji/white-heavy-check-mark.png} ($\slsp$) & $\testlayout_i$ &  500 & 0.2 $\rightarrow$ 0.005 & 250 ep\\
    \hline
    $\BRa(H)$ & \includegraphics[width=6pt]{images/emoji/white-heavy-check-mark.png} ($\slsp$) & $\testlayout_i$ &  500 & 0.2 $\rightarrow$ 0.005 & 250 ep\\
    \hline
    \end{tabular}
    \vskip 0.6em
    \caption{All best response conditions. Note that we also provide $\acr$ to all best response agents, and give equal access to training for each group of conditions.}
 \label{tab:br-summary}
\end{table}


\section{Additional analysis for human modeling conditions}\label{sup:human-modeling}

In the human data used for our simulated humans' parameter fitting, roughly 40\%-65\% of the actions were ``wait'' actions. This is likely due to the rapid clock speed of the game, making it difficult for humans to perform constructive actions at every timestep. 

\begin{figure}[h]
\centering
\begin{subfigure}[b]{0.49\textwidth}
   \includegraphics[width=\linewidth]{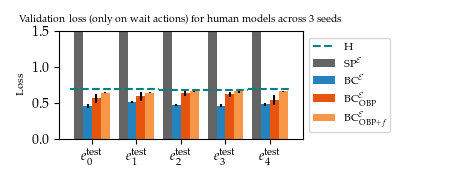}
   \label{fig:wait only} 
\end{subfigure}
\begin{subfigure}[b]{0.49\textwidth}
   \includegraphics[width=\linewidth]{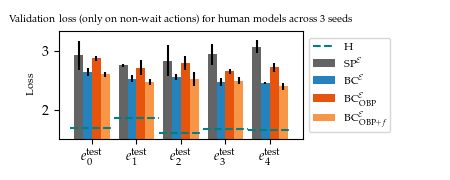}
   \label{fig:no-waiting} 
\end{subfigure}
\vskip -1.5em
\caption{\textbf{Breaking down the loss}: Even though $\mlbc$, $\mlbcmlsp$, and $\mlbcmlspfreeze$ achieve similar loss when averaged across all actions, $\acr$ helps prevent human models from overpredicting the probability of waiting, as $\mlbc$ does.}
\label{fig:wait-info}
\end{figure}

In looking at \Cref{fig:val-comp-bc} (right) in the main text, we were surprised that $\acr$-methods performed as well as non-$\acr$ methods in terms of validation loss. To further investigate the cause of this similar performance, we broke down training datapoints into two groups: those where in which the ground truth action was or wasn't a "wait" action. We show this in \Cref{fig:wait-info}.

We discover that $\mlbc$ assigns high probability on wait actions (thus achieving even lower loss than ground truth human H does). This supports our intuition that without having a prior, $\mlbc$ will try to find the easiest route to lower loss: overemphasising ``wait'' action predictions. On the other hand, by using $\acr$, $\mlbcmlsp$ and $\mlbcmlspfreeze$ both achieve more human-like loss on wait actions. We speculate that this is what makes $\mlbc$ a \textit{qualitatively} worse human model which is  incapable of achieving reward, as mentioned in the main text.


\end{document}